\documentclass[sigconf]{acmart}

\usepackage{booktabs} 
\usepackage{dsfont}
\usepackage{stmaryrd}
\usepackage{color}
\usepackage{bm}
\usepackage{CJK}
\usepackage{subfigure}
\usepackage{multirow}
\usepackage{algorithm}
\usepackage{algorithmic}
\usepackage{array}
\usepackage{subfigure}
\usepackage{tikz}
\usepackage{pgfplots}
\usepackage{epstopdf}
\pgfplotsset{compat=1.14}
\usetikzlibrary{patterns}
\usepackage{footnote}
\makesavenoteenv{tabular}
\makesavenoteenv{table}
\newcommand{\PreserveBackslash}[1]{\let\temp=\\#1\let\\=\temp}
\newcolumntype{C}[1]{>{\PreserveBackslash\centering}p{#1}}
\newcolumntype{R}[1]{>{\PreserveBackslash\raggedleft}p{#1}}
\newcolumntype{L}[1]{>{\PreserveBackslash\raggedright}p{#1}}
\setcopyright{none}

\begin{document}

\title{A Tree Search algorithm For Sequence Labeling}

\author{Yadi Lao, Jun Xu$^*$, Yanyan Lan, Jiafeng Guo, Sheng Gao, Xueqi Cheng, Jun Guo}

\affiliation{%
Beijing University of Posts and Telecommunications\\%
CAS Key Lab of Network Data Science and Technology\\%
Institute of Computing Technology, Chinese Academy of Science\\%
}
 
\begin{abstract}
In this paper we propose a novel reinforcement learning based model for sequence tagging, referred to as MM-Tag. Inspired by the success and methodology of the AlphaGo Zero, MM-Tag formalizes the problem of sequence tagging with a Monte Carlo tree search (MCTS) enhanced Markov decision process (MDP) model, in which the time steps correspond to the positions of words in a sentence from left to right, and each action corresponds to assign a tag to a word. Two long short-term memory networks (LSTM) are used to summarize the past tag assignments and words in the sentence. Based on the outputs of LSTMs, the policy for guiding the tag assignment and the value for predicting the whole tagging accuracy of the whole sentence are produced. The policy and value are then strengthened with MCTS, which takes the produced raw policy and value as inputs, simulates and evaluates the possible tag assignments at the subsequent positions, and outputs a better search policy for assigning tags. A reinforcement learning algorithm is proposed to train the model parameters. Our work is the first to apply the MCTS enhanced MDP model to the sequence tagging task. We show that MM-Tag can accurately predict the tags thanks to the exploratory decision making mechanism introduced by MCTS. Experimental results show based on a chunking benchmark showed that MM-Tag outperformed the state-of-the-art sequence tagging baselines including CRF and CRF with LSTM.  
\end{abstract}

\maketitle

\section{Introduction}\label{sec:Intro}
Sequence tagging, including POS tagging, chunking, and name entity recognition, has gained considerable research attention for a few decades. Using the chunking as an example, given a sentences of text (e.g., a sentence), each of the word in a sequence receives a ``tag'' (class label) that expresses its phrase type.

Existing sequence tagging models are be categorized into the statistical models and the deep neural networks based models. Traditional research on sequence tagging focus on the linear statistical models, including the maximum entropy (ME) classifier~\cite{ratnaparkhi1996maximum} and maximum entropy Markov models (MEMMs)~\cite{mccallum2000maximum}. These models predict a distribution of tags for each time step and then use beam-like decoding to find optimal tag sequences. Lafferty et al. proposed conditional random fields (CRF)~\cite{lafferty2001conditional} to leverage global sentence level feature and solve the label bias problem in MEMM. All the linear statistical models rely heavily on hand-crafted features, e.g., the word spelling features for the task of part-of-speech. Motivated by the success of deep learning, deep neural networks based models have been proposed for sequence tagging in recent years. Most of them directly combine the deep neural networks with CRF. For example, Huang~\cite{huang2015bidirectional} used a bidirectional LSTM to automatically extract word-level representations and then combined with CRF for jointly label decoding. Ma~\cite{ma2016end} introduced a neural network architecture that both word level and character level features are used, in which bidirectional LSTM, CNN, and CRF are combined. In recent years, reinforcement learning is also proposed for the task. For example, Maes et al. formalized the sequence tagging task as a Markov decision process (MDP) and used the reinforcement learning algorithm SARSA to construct optimal sequence directly in a greedy manner~\cite{maes2007sequence}.  Feng et al. proposed a novel model  to address the noise problem of relation classification task caused by distant supervision, in which an instance selector designed with REINFORCE algorithm is used to assign select or delete action (label) for every sentence \citet{feng2018reinforcement}.


Inspired by the reinforcement learning model of AlphaGO~\cite{silver2016mastering} and AlphaGO Zero~\cite{silver2017mastering} programs designed for the Game of Go, in this paper we propose to solve the sequence tagging with a Monte Carlo tree search (MCTS) enhanced Markov decision process (MDP). The new sequence tagging model, referred to as MM-Tag (MCTS enhanced MDP for Tagging), makes use of an MDP to model the sequential tag assignment process in sequence tagging. At each position (corresponding to a ranking position), based on the past words and tags, two long short-term memory networks (LSTM) are used to summarize the past words and tags, respectively. Based on the outputs of these two LSTMs, a policy function (a distribution over the valid tags) for guiding the tag assignment and a value function for estimating the accuracy of tagging are produced. To avoid the problem of assigning tags without utilizing the whole sentence level tags, in stead of choosing a tag directly with the raw policy predicted by the policy function, MM-Tag explores more possibilities in the whole space. The exploration is conducted with the MCTS guided by the produced policy function and value function, resulting a strengthened search policy for the tag assignment. Moving to the next iteration, the algorithm moves to the next position and continue the above process until at the end of the sentence. 

Reinforcement learning is used to train the model parameters. In the training phase, at each learning iteration and for each training sentence (and the corresponding labels), the algorithm first conducts an MCTS inside the training loop, guided by the current policy function and value function. Then the model parameters are adjusted to minimize the loss function. The loss function consists of two terms: 1) the squared error between the predicted value and the final ground truth accuracy of the whole sentence tagging; and 2) the cross entropy of the predicted policy and the search probabilities for tags selection. Stochastic gradient descent is utilized for conducting the optimization. 

To evaluate the effectiveness of MM-Tag, we conducted experiments on the basis of CoNLL 2000 chunking dataset. The experimental results showed that MM-Tag can significantly outperform the state-of-the-art sequence tagging approaches, including the linear statistical models of CRF and neural network-based models of BI-LSTM-CRF. We analyzed the results and showed that MM-Tag improved the performances through conducting lookahead MCTS to explore in the whole tagging space.



\section{\mbox{MDP formulation of sequence tagging}}
In this section, we introduce the proposed MM-Tag model. 
\subsection{Sequence tagging as an MDP} 
Suppose that $\mathbf{X}=\{\mathbf{x}_1, \cdots, \mathbf{x}_M\}$ is a sequence of words (sentence) to be labeled, and $\mathbf{Y}=\{y_1, \cdots, y_M\}$ is the corresponding ground truth tag sequence. All components $\mathbf{x}_i$ of $\mathbf{X}$ are the $L$-dimensional preliminary representations of the words, i.e., the word embedding. All components $y_i$ of $\mathbf{Y}$ are assumed to be selected from a finite tag set $\mathcal{Y}$. From example, $\mathcal{Y}$ may be the set of possible part-of-speech tags. The goal of sequence tagging is to construct a model that can automatically assign a tag to each word in the input sentence $\mathbf{X}$.

MM-Tag formulates the assignment of tags to sentences as a process of sequential decision making with an MDP in which each time step corresponds to a position in the sentence. The states, actions, transition function, value function, and policy function of the MDP are set as:

\textbf{States $S$:} We design the state at time step $t$ as a pair $s_t =[\mathbf{X}_t = \{\mathbf{x}_1, \cdots, \mathbf{x}_t\}, \mathbf{Y}_t =\{ \mathbf{y}_1, \cdots, \mathbf{y}_{t-1}\}]$, where $\mathbf{X}_t$ is the preliminary representation of the prefix of the input sentence of length $t$. $\mathbf{Y}_t$ is the prefix of the label sequence of length $t-1$. At the beginning ($t=1$), the state is initialized as  $s_1 =[\{\mathbf{x}_1\}, \{\emptyset\}]$, where $\emptyset$ is the empty sequence. 
      
\textbf{Actions $\mathcal{A}$:} At each time step $t$, the $\mathcal{A}(s_t)\subseteq \mathcal{Y}$ is the set of actions the agent can choose. That is, the action $a_t\in\mathcal{A}(s_t)$ actually chooses a tag $y_t \in \mathcal{Y}$ for word $\mathbf{x}_t$.

\textbf{Transition function $T$:} $T:S\times \mathcal{A}\rightarrow S$ is defined as 
$$
s_{t+1}=T(s_t, a_t)  = T([\mathbf{X}_t, \mathbf{Y}_t], a_t) = [\mathbf{X}_t \oplus \{\mathbf{x}_{t+1}\}, \mathbf{Y}_t \oplus \{a_t\}]
$$
where $\oplus$ appends $\mathbf{x}_{t+1}$ and $a_t$ to $\mathbf{X}_t$ and $\mathcal{Y}_t$, respectively. At each time step $t$, based on state $s_t$ the system chooses an action (tag) $a_t$ for the word position $t$. Then, the system moves to time step $t+1$ and the system transits to a new state $s_{t+1}$: first, the word sequence $\mathbf{X}_t$ is updated by concatenating the next word $\mathbf{x}_{t+1}$; second, the system appends the selected tag to the end of $\mathbf{Y}_t$, generating a new tag sequence. 

\textbf{Value function $V$:} The state value function $V: S\rightarrow \mathbb{R}$ is a scalar evaluation, predicting the accuracy of the tag assignments for the whole sentence (an episode), on the basis of the input state. The value function is learned so as to fit the real tag assignment accuracies of the training sentences. 

In this paper, we use two LSTMs to respectively map the word sequence $\mathbf{X}_t$ and tag sequence $\mathbf{Y}_t$ in the state $s_t$ to two real vectors, and then define the value function as nonlinear transformation of the weighted sum of
the LSTM's outputs:
\begin{equation}\label{eq:valueFunction}
V(s) = \sigma\left(\left\langle \mathbf{w},\mathbf{g}(s) \right\rangle + b_v\right)	
\end{equation}
where $\mathbf{w}$ and $b_v$ are the weight vector and the bias to be learned during training, $\sigma(x) = \frac{1}{1+e^{-x}}$ is the nonlinear sigmoid function, and $\mathbf{g}(s)$ is a concatenation of the outputs from the word LSTM LSTM$_X$ and tag LSTM LSTM$_Y$: 
\begin{equation}\label{eq:state}
\mathbf{g}(s) = \left[\mathrm{LSTM}_X(\mathbf{X}_t)^T, \mathrm{LSTM}_Y(\mathbf{Y}_{t-1})^T\right]^T. 
\end{equation}

The two LSTM networks are defined as follows: given $s=[\mathbf{X}_t=\{\mathbf{x}_1, \cdots, \mathbf{x}_t\}, \mathbf{Y}_t=\{\mathbf{y}_1, \cdots, \mathbf{y}_{t-1}\}]$, where $\mathbf{x}_k (k=1,\cdots, t)$ is the word at $k$-th position, represented with its word embedding. $\mathbf{y}_k (k=1,\cdots, T-1)$ is the label at $k$-th position, represented with one hot vector. LSTM$_X$ outputs a representation $h_k$ for position $k$:
\begin{align}
\nonumber \mathbf{f}_k = & \sigma(\mathbf{W}_{f}^X\mathbf{x}_k + \mathbf{U}_{f}^X\mathbf{h}_{k-1}+\mathbf{b}_f^X), \mathbf{i}_k =  \sigma(\mathbf{W}_{i}^X\mathbf{x}_k + \mathbf{U}_{i}^X\mathbf{h}_{k-1}+\mathbf{b}_i^X),\\
\nonumber	\mathbf{o}_k = & \sigma(\mathbf{W}_{o}^X\mathbf{x}_k+\mathbf{U}_{o}^X\mathbf{h}_{k-1}+\mathbf{b}_o^X),\\
\nonumber	\mathbf{c}_k = & \mathbf{f}_{k}\circ \mathbf{c}_{k-1}+\mathbf{i}_k \circ \tanh(\mathbf{W}_{c}^X\mathbf{x}_k + \mathbf{U}_{c}^X\mathbf{h}_{k-1} +\mathbf{b}_c^X),\\	
\nonumber	\mathbf{h}_k = & \mathbf{o}_k \circ \tanh(\mathbf{c}_k),
\end{align}where $\mathbf{h}$ and $\mathbf{c}$ are initialized with zero vector; operator ``$\circ$'' denotes the element-wise product and ``$\sigma$'' is applied to each of the entries; the variables $\mathbf{f}_k, \mathbf{i}_k, \mathbf{o}_k, \mathbf{c}_k$ and $\mathbf{h}_k$ denote the forget gate's activation vector, input gate's activation vector, output gate's activation vector, cell state vector, and output vector of the LSTM block, respectively. $\mathbf{W}_{f}^X,\mathbf{W}_{i}^X,\mathbf{W}_{o}^X, \mathbf{U}_{f}^X,\mathbf{U}_{i}^X,\mathbf{U}_{o}^X, \mathbf{b}_{f}^X,\mathbf{b}_{i}^X,\mathbf{b}_{o}^X$ are weight matrices and bias vectors need to be learned during training. The output vector and cell state vector at the $t$-th cell are concatenated as the output of LSTM, that is 
$$
\textrm{LSTM}_X(\mathbf{X}_t) = \left[{\mathbf{h}_t}^T, {\mathbf{c}_t}^T\right]^T.
$$

The function $\textrm{LSTM}_Y(\mathbf{Y}_{t-1})$, which used to map the tag sequence $\mathbf{Y}_{t-1}$ into a real vector, is defined similarly to that of for LSTM$_X$. 

\textbf{Policy function $\mathbf{p}$:} The policy $\mathbf{p}(s)$ defines a function that takes the state as input and output a distribution over all of the possible actions $a\in\mathcal{A}(s)$. Specifically, each probability in the distribution is a normalized soft-max function whose input is the bilinear product of the state representation in Equation~(\ref{eq:state}) and the selected tag:
\[
p(a|s) = \frac{\exp\left\{\Phi(a)^T \mathbf{U}_p ~\mathbf{g}(s)\right\}}{\sum_{a'\in\mathcal{A}(s)} \exp\left\{\Phi(a')^T \mathbf{U}_p ~\mathbf{g}(s)\right\}},
\]
where $\Phi(a)$ is the one hot vector for representing the tag $a$ and $\mathbf{U}_p$ is the parameter in bilinear product. The policy function $\mathbf{p}(s)$ is:  
\begin{equation}\label{eq:Policy}
\mathbf{p}(s) = \langle p(a_1|s), \cdots, p(a_{|\mathcal{A}(s)|}|s)\rangle.
\end{equation}

\subsection{Strengthening raw policy with MCTS}
Tagging directly with the predicted raw policy $\mathbf{p}$ in Equation~(\ref{eq:Policy}) may lead to suboptimal results because the policy is calculated based on the past tags. The raw policy has no idea about the tags that will be assigned for the future words. To alleviate the problem, following the practices in AlphaGo~\cite{silver2016mastering} and AlphaGo Zero~\cite{silver2017mastering}, we propose to conduct lookahead search with MCTS. That is, at each position $t$, an MCTS search is executed, guided by the policy function $\mathbf{p}$ and the value function $V$, and output a strengthened new search policy $\bm{\pi}$. Usually, the search policy $\bm{\pi}$ has high probability to select a tag with higher accuracy than the raw policy $\mathbf{p}$ defined in Equation~(\ref{eq:Policy}).

Algorithm~\ref{alg:TreeSearch} shows the details of the MCTS in which each tree node corresponds to an MDP state. It takes a root node $s_R$, value function $V$ and policy function $\mathbf{p}$ as inputs. The algorithm iterates $K$ times and outputs a strengthened search policy $\bm{\pi}$ for selecting a tag for the root node $s_R$. Suppose that each edge $e(s, a)$ (the edge from state $s$ to the state $T(s, a)$) of the MCTS tree stores an action value $Q(s, a)$, visit count $N(s, a)$, and prior probability $P(s, a)$. At each of the iteration, the MCTS executes the following steps:

\textbf{Selection}: Each iterations starts from the root state $s_R$ and iteratively selects the documents that maximize an upper confidence bound:
\begin{equation}\label{eq:Selection}
	a_t = \arg\max_a (Q(s_t, a) + \lambda U(s_t, a)),
\end{equation}
where $\lambda >0$ is the tradeoff coefficient, and the bonus $U(s_t, a) =  p(a|s_t)\frac{\sqrt[]{\sum_{a'\in\mathcal{A}(s_t)} N(s_t, a')}}{1 + N(s_t, a)}$. $U(s_t, a)$ is proportional to the prior probability but decays with repeated visits to encourage exploration.

\textbf{Evaluation and expansion}: When the traversal reaches a leaf node $s_L$, the node is evaluated with the value function $V(s_L)$ (Equation~(\ref{eq:valueFunction})). Note following the practices in AlphaGo Zero, we use the value function instead of rollouts for evaluating a node. 

Then, the leaf node $s_L$ may be expanded. Each edge from the leaf position $s_L$ (corresponds to each action $a\in\mathcal{A}(s_L)$) is initialized as: $P(s_L, a) = p(a|s_L)$ (Equation~(\ref{eq:Policy})), $Q(s_L, a)= 0$, and $N(s_L, a) = 0$. In this paper all of the available actions of $s_L$ are expanded.

\textbf{Back-propagation and update}: At the end of evaluation, the action values and visit counts of all traversed edges are updated. For each edge $e(s, a)$, the prior probability $P(s, a)$ is kept unchanged, and $Q(s,a)$ and $N(s, a)$ are updated: 
\begin{equation}\label{eq:UpdateQN}
	Q(s, a) \leftarrow  \frac{Q(s, a) \times N(s, a) + V(s_L)}{N(s, a) + 1}; N(s, a) \leftarrow  N(s, a) + 1.
\end{equation}
\textbf{Calculate the strengthened search policy}: Finally after iterating $K$ times, the strengthened search policy $\bm\pi$ for the root node $s_R$ can be calculated according to the visit counts $N(s_R, a)$ of the edges starting from $s_R$:
\begin{equation}\label{eq:SearchProb}
\pi(a|s_R) = \frac{N(s_R, a)}{\sum_{a'\in\mathcal{A}(s_R)} N(s_R, a')},
\end{equation}
for all $a\in\mathcal{A}(s_R)$.

\begin{algorithm}[t]
\caption{TreeSearch}\label{alg:TreeSearch}
\renewcommand{\algorithmicrequire}{\textbf{Input:}}
\renewcommand{\algorithmicensure}{\textbf{Output:}}
\begin{algorithmic}[1]
\REQUIRE root $s_R$, value $V$, policy $\mathbf{p}$, search times $K$
\ENSURE Search policy $\bm{\pi}$
\FOR{$k = 0$ \TO $K-1$}
\STATE $s_L\leftarrow s_R$
\STATE \COMMENT{Selection}
\WHILE{$s_L \textrm{ is not a leaf node}$}
	\STATE $a \leftarrow \arg\max_{a\in\mathcal{A}(s_L)} Q(s_L, a) +\lambda \cdot U(s_L, a)$\COMMENT{Eq.~(\ref{eq:Selection})}
	\STATE $s_L \leftarrow \textrm{ child node pointed by edge }(s_L, a)$ 
\ENDWHILE

\STATE \COMMENT{Evaluation and expansion}
    \STATE $v\leftarrow V(s_L)$ \COMMENT{simulate $v$ with value function $V$}
	\FORALL{$a \in\mathcal{A}(s_L)$} 
		\STATE Expand an edge $e$ to node $ s = [s_L.\mathbf{X}_t, s_L.\mathbf{Y}_t\oplus \{a\}]$
		\STATE $e.P \leftarrow p(a|s_L); e.Q \leftarrow 0; e.N \leftarrow  0$\COMMENT{init edge properties}
	\ENDFOR
\STATE \COMMENT{Back-propagation}
\WHILE{$s_L \neq s_R$} 
	\STATE $s \leftarrow \textrm{ parent of }s_L; e \leftarrow \textrm{ edge from }s\textrm{ to }s_L$
	\STATE $e.Q \leftarrow \frac{e.Q\times e.N + v}{e.N + 1}$\COMMENT{Eq.~(\ref{eq:UpdateQN})}
	\STATE $e.N \leftarrow e.N + 1; s_L \leftarrow s$
\ENDWHILE
\ENDFOR

\STATE \COMMENT{Calculate tree search policy. Eq.~(\ref{eq:SearchProb})}
\FORALL{$a\in\mathcal{A}(s_R)$}
	\STATE $\pi(a|s_R)\leftarrow \frac{e(s_R, a).N}{\sum_{a'\in\mathcal{A}(s_R)}e(s_R, a').N}$
\ENDFOR
\RETURN $\bm{\pi}$
\end{algorithmic}
\end{algorithm}

\subsection{Learning and inference algorithms}
\subsubsection{Reinforcement learning of the parameters}
The model has parameters $\mathbf{\Theta}$ (including $\mathbf{w}, b_v, \mathbf{U}_p$, and parameters in LSTM$_X$ and LSTM$_Y$) to learn. In the training phase, suppose we are given $N$ labeled sentence $D = \{ (\mathbf{X}^{(n)}, \mathbf{Y}^{(n)})\}_{n=1}^{N}$. Algorithm~\ref{alg:Train} shows the training procedure. First, the parameters $\mathbf\Theta$ is initialized to random weights in $[-1, 1]$. At each subsequent iteration, for each tagged sentence $(\mathbf{X}, \mathbf{Y})$, a tag sequence is predicted for $\mathbf{X}$ with current parameter setting: at each position $t$, an MCTS search is executed, using previous iteration of value function and policy function, and a tag $a_t$ is selected according to the search policy $\bm{\pi}_t$. The ranking terminates at the end of the sentence and achieved a predicted tag sequence $(a_1, \cdots, a_M)$. Given the ground truth tag sequence $\mathbf{Y}$, the overall prediction accuracy of the sentence $\mathbf{X}$ is calculated, denoted as $r$. The data generated at each time step $E=\{(s_t, \bm{\pi}_t)\}_{t=1}^M$ and the final evaluation $r$ are utilized as the signals in training for adjusting the value function. The model parameters are adjusted to minimize the error between the predicted value $V(s_t)$ and the whole sentence accuracy $r$, and to maximize the similarity of the policy $\mathbf{p}(s_t)$ to the search probabilities $\bm{\pi}_t$. Specifically, the parameters $\mathbf\Theta$ are adjusted by gradient descent on a loss function $\ell$ that sums over the mean-squared error and cross-entropy losses, respectively:
\begin{equation}\label{eq:loss}
	\ell(E, r) = \sum_{t=1}^{|E|} \left(\left(V(s_t) - r\right)^2 +\sum_{a\in\mathcal{A}(s_t)}\pi_t(a|s_t) \log \frac{1}{p(a|s_t)}\right).
\end{equation}
The model parameters are trained by back propagation and stochastic gradient descent. Specifically, we use AdaGrad~\cite{duchi2011adaptive} on all parameters in the training process. 

\begin{algorithm}[t]
\caption{Train MM-Tag model}\label{alg:Train}
\renewcommand{\algorithmicrequire}{\textbf{Input:}}
\renewcommand{\algorithmicensure}{\textbf{Output:}}
\begin{algorithmic}[1]
\REQUIRE Labeled data $D=\{ (\mathbf{X}^{(n)}, \mathbf{Y}^{(n)})\}_{n=1}^N$, learning rate $\eta$, number of search $K$
\ENSURE $\mathbf{\Theta}$
\STATE \text{Initialize} $\mathbf{\Theta} \leftarrow$ random values in $[-1, 1]$
\REPEAT
	\FORALL{$(\mathbf{X}, \mathbf{Y})\in D$}
		\STATE {$s \leftarrow [\mathbf{x}_1, \emptyset]; M \leftarrow |X|; E\leftarrow \emptyset$ }
		\FOR{$t=1$ \TO $M$}
            \STATE $\bm{\pi} \leftarrow \mathrm{TreeSearch}(s, V, \mathbf{p}, K)$ \COMMENT{Alg.~(\ref{alg:TreeSearch})}
			\STATE $a =\arg\max_{a\in\mathcal{A}(s)} \pi(a|s)$ \COMMENT{select the best tag}
			\STATE $E\leftarrow E \oplus \{(s, \bm{\pi})\}$
			\STATE $s \leftarrow [s.\mathbf{X}_t \oplus \{\mathbf{x}_{t+1}\}, s.\mathbf{Y}_t \oplus \{a\}]$
		\ENDFOR
		\STATE $r \leftarrow \textrm{Acccuracy}(\mathbf{Y}, s.\mathbf{Y}_M)$\COMMENT{overall accuracy}
		\STATE $\mathbf\Theta \leftarrow \mathbf\Theta -\eta \frac{\partial \ell(E, r)}{\partial \mathbf\Theta}$ \COMMENT{$\ell$ is defined in Eq.~(\ref{eq:loss})}
	\ENDFOR
\UNTIL {converge}
\RETURN $\mathbf{\Theta}$
\end{algorithmic}
\end{algorithm}

\subsubsection{Inference}
The inference of the tag sequence for a sentence is shown in Algorithm~\ref{alg:RLRank_MCTS}. Given a sentence $\mathbf{X}$, the system state is initialized as $s_1=[\{\mathbf{x}_1\}, \emptyset]$. Then, at each of the time steps $t=1,\cdots, M$, the agent receives the state $s_t=[\mathbf{X}_t, \mathbf{Y}_t]$ and search the policy $\bm\pi$ with MCTS, on the basis of the value function $V$ and policy function $\mathbf{p}$. Then, it chooses an action $a$ for the word at position $t$. Moving to the next iteration ${t+1}$, the state becomes $s_{t+1}=[\mathbf{X}_{t+1}, \mathbf{Y}_{t+1}]$. The process is repeated until the end of the sentence is reached. 

We implemented the MM-Tag model based on TensorFlow and the code can be found at the Github repository \url{http://hide_for_anonymous_review}. 
\begin{algorithm}[t]
\caption{MM-Tag Inference}\label{alg:RLRank_MCTS}
\renewcommand{\algorithmicrequire}{\textbf{Input:}}
\renewcommand{\algorithmicensure}{\textbf{Output:}}
\begin{algorithmic}[1]
\REQUIRE sentence $\mathbf{X}=\{\mathbf{x}_1, \cdots, \mathbf{x}_M\}$, value function $V$, policy function $\mathbf{p}$, and search times $K$,
\ENSURE label sequence $\mathbf{Y}$
\STATE $s \leftarrow [\{\mathbf{x}_1\}, \emptyset]; M \leftarrow |\mathbf{X}|$
\FOR{$t=1$ \TO $M$}
	\STATE $\bm{\pi} \leftarrow \mathrm{TreeSearch}(s, V, \mathbf{p}, K)$
	\STATE $a \leftarrow \arg\max_{a\in\mathcal{A}(s)} \pi(a|s)$ 
	\STATE $s \leftarrow [s.\mathbf{X} \oplus \{\mathbf{x}_{t+1}\}, s.\mathbf{Y}\oplus \{a\}]$
\ENDFOR
\RETURN $s.\mathbf{Y}$
\end{algorithmic}
\end{algorithm}

\section{Experiments}

We tested the performances of MM-Tag on subsets of CoNLL 2000 chunking set\footnote{https://www.clips.uantwerpen.be/conll2000/chunking/}. In chunking task, each word is tagged with its phrase type, e.g., tag ``B-NP'' indicates a word starting a noun phrase.
Considering MM-Tag is time consuming for parsing the long sentences, we constructed a short sentence subset which was randomly selected from the whole CoNLL 2000 chunking set and the sentences longer than 13 words were removed. The final short sentence subset consists of 1000 sentences and the average sentence length is 9 words. Among them, 900 were used for training and 100 were used for testing. 
All of the words in the sentences were represented with the word embeddings. In the experiments, we used the publicly available GloVe 100-dimensional embeddings trained on 6 billion words from Wikipedia and Gigaword~\cite{pennington2014glove}.

We compare MM-Tag with linear statistical models of CRF implemented with an open software CRFsuite\footnote{http://www.chokkan.org/software/crfsuite/} ~\cite{okazaki2007crfsuite} and neural models of LSTM-CRF and BI-LSTM-CRF, following the configurations in~\cite{huang2015bidirectional}. For CRF, 935 
spelling features and context features were extracted. 
The features including word identity, word suffix, word shape, word POS tag from current and nearby words etc.  

For MM-Tag, the number of search times $K$, the learning rate $\eta$, the tree search trade-off parameter $\lambda$, and the number of hidden units in LSTM $h$ were set to  $K = 4000$, $\eta=0.001$, $\lambda=0.25$, and $h=200$.

Table~\ref{tab:result} reports the performances of MM-Tag and baseline methods in terms of tagging precision, recall, F1, and accuracy. Boldface indicates the highest scores among all runs. From the result we can see that, MM-Tag outperformed all of the baseline methods in terms of all of the evaluation metrics, indicating the effectiveness of the proposed MM-Tag model. 
We note that neural methods (LSTM+CRF and BiLSTM+CRF) were underperformed the CRF and MM-Tag. The reason may because the short sentence subset is not sufficient enough to learn the large amount of parameters in neural networks. 


\begin{table}
\centering
\caption{Performance comparison of all methods.}\label{tab:result}
\begin{tabular}{c|cccc}
\hline                                     &Precision                & Recall     & F1        & Accuracy\\\hline
\hline
CRF                                     &   93.98\%		      & 94.47\%       & 94.13\%             & 94.54\%\\ \hline
LSTM+CRF                       &  88.04\%	           &88.43\%       & 88.17\%             &89.99\%  \\ \hline
BiLSTM+CRF                   &   89.57\%		      & 89.81\%       & 89.65\%             & 90.61\%\\ \hline
MM-Tag                           &   \textbf{95.75\%}         &\textbf{95.47\%}       &\textbf{ 94.82\% }            &\textbf{95.77\%}  \\ \hline
\end{tabular}
\end{table}


\section{Conclusion}
In this paper we have proposed a novel approach to sequence tagging, referred to as MM-Tag. MM-Tag formalizes the tagging of a sentence as a sequence of decision-making with MDP. The lookahead MCTS is used to strengthen the raw predicted policy so that the search policy has high probability to select the correct for each word. Reinforcement learning is utilized to train the model parameters. MM-Tag enjoys several advantages: tagging with the shared policy and the value functions, end-to-end learning, and high accuracy in tagging. Experimental results show that MM-Tag outperformed the baselines of CRF, LSTM-CRF, and BI-LSTM-CRF.

%
\bibliographystyle{ACM-Reference-Format}
\bibliography{mmtag}  

%
%
\end{document}